\documentclass{llncs}
\usepackage{float}
\usepackage{tabularx,ragged2e}
\usepackage{amsmath}
\usepackage{booktabs}
\usepackage{multirow}
\usepackage{subfigure}
\usepackage{lscape}
\usepackage[table,xcdraw]{xcolor}
\usepackage{makeidx}  % allows for indexgeneration
 \usepackage{multirow}
 \usepackage[table,xcdraw]{xcolor}
\usepackage{amssymb}
\usepackage{array}
\newcolumntype{P}[1]{>{\centering\arraybackslash}p{#1}}
\usepackage{graphicx}
\usepackage{tabu}
\usepackage[flushleft]{threeparttable}
\usepackage{url}
\makeindex
\begin{document}

\title{Multi-Class Lesion Diagnosis with Pixel-wise Classification Network}

\author{Manu Goyal \inst{1} \and Jiahua Ng \inst{2} \and Moi Hoon Yap \inst{1}}

\index{Goyal, Manu}
\index{Yap, Moi Hoon}
% use the command \index{<name>} for index entries

\institute{Visual Computing Lab, Manchester Metropolitan University, M1 5GD, UK \\
\and
University of Sheffield, S10 2TN, UK}

\maketitle

%\clearpage

\begin{abstract}
	 Lesion diagnosis of skin lesions is a very challenging task due to high inter-class similarities and intra-class variations in terms of color, size, site and appearance among different skin lesions. With the emergence of computer vision especially deep learning algorithms, lesion diagnosis is made possible using these algorithms trained on dermoscopic images. Usually, deep classification networks are used for the lesion diagnosis to determine different types of skin lesions. In this work, we used pixel-wise classification network to provide lesion diagnosis rather than classification network. We propose to use DeeplabV3+ for multi-class lesion diagnosis in dermoscopic images of Task 3 of \textit{ISIC Challenge 2018}.  We used the various post-processing methods with DeeplabV3+ to determine the lesion diagnosis in this challenge and submitted the test results. 
	% and submitted the various ensemble techniques using both networks to produce final masks for ISIC-2018 testing set. 
	%	Early detection of skin cancer, particularly melanoma, is crucial to enable advanced treatment. Due to the rapid growth of skin cancers, there is a growing need of computerized analysis for skin lesions. These processes including detection, classification, and segmentation. The state-of-the-art public available datasets for skin lesions are often accompanied with very limited amount of segmentation ground truth labelling as it is laborious and expensive.  In this work, 
\end{abstract}

\section{Introduction}
	 Automated skin lesions analysis such as skin cancer \cite{goyal2017multi} and diabetic foot ulcers \cite{goyal2017fully} has gain its popularity, thanks to the advancement of computer vision, machine learning and deep learning algorithms. Skin cancer is the most common cancer among all other cancers \cite{pathan2018techniques}. The malignant skin lesions consist of the melanocytic lesion, i.e. melanoma, and non-melanocytic lesion, i.e. basal cell carcinoma. Although melanoma is the least common type of skin cancer, it is the most aggressive and deadly cancer \cite{Seer2017}. Hence, it is important to have early detection to save life. According to the prediction of Melanoma Foundation \cite{melanomafoundation2017}, the estimated new cases of melanoma in the United States is 87,110 (200\% increased since 1973) with 9,730 predicted deaths. For \textit{International Skin Imaging Collaboration (ISIC) 2018 Challenge}, we propose the use of DeeplabV3+ for \textit{Task 3: lesion diagnosis} and we describe our methods in the following section.
	%In current practice, medical experts primarily examine and assess the skin lesion of patients on visual inspection with manual measurements tools. With limited healthcare resources, periodic check-ups of doctors are not possible for each patient for detection of abnormal skin lesions. The computer vision along with machine algorithms have a great potential to provide the complete tele-medicine systems on mobile devices which can detect these abnormal skin lesions.

\section{Methodology}
This section discusses the challenge dataset, the preparation of ground truth and proposed methods.

\subsection{ISIC 2018 Challenge Dataset for Task 3}

\textit{ISIC 2018 Lesion Diagnosis Challenge} consists of 10015 training images, 193 validation images and 1512 testing images \cite{codella2017skin}. It consists of 7 different types of skin lesions including melanoma, nevi, basal cell carcinoma, actinic keratosis, benign keratosis, dermatofibroma, vascular lesion. Benign keratosis is super category of different types of keratosis lesions such as seborrheic keratosis, solar lentigo and lichen planus-like keratosis. The dataset used by the challenge is also known as, \textit{HAM10000} which is a collection of 11,788 dermoscopic images of different skin lesions collected from the multiple sources around the world \cite{tschandl2018ham10000}. We did not use any external data.  To improve the performance and reduce the computational costs, we resized all the images to 500 $\times$ 375.   

\subsection{Preprocessing and Preparation of Ground Truth}
\begin{figure}
	\centering
	\begin{tabular}{ccc}
		\includegraphics[width=3cm,height=2.2cm]{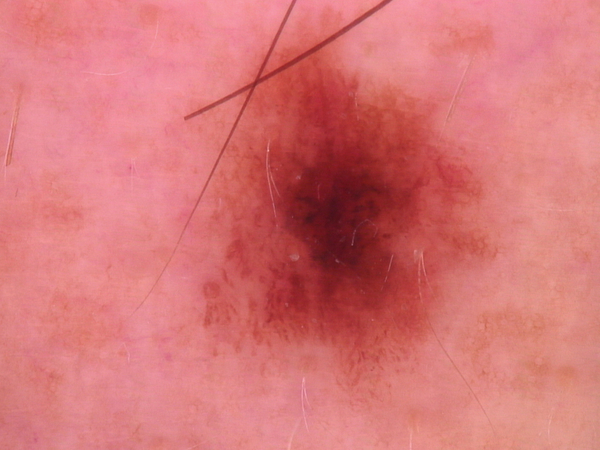} &
		\includegraphics[width=3cm,height=2.2cm]{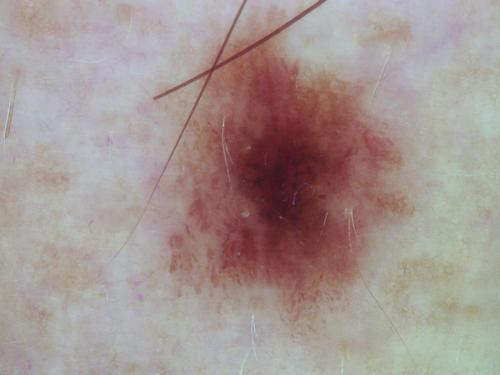}    
		\\
		\includegraphics[width=3cm,height=2.2cm]{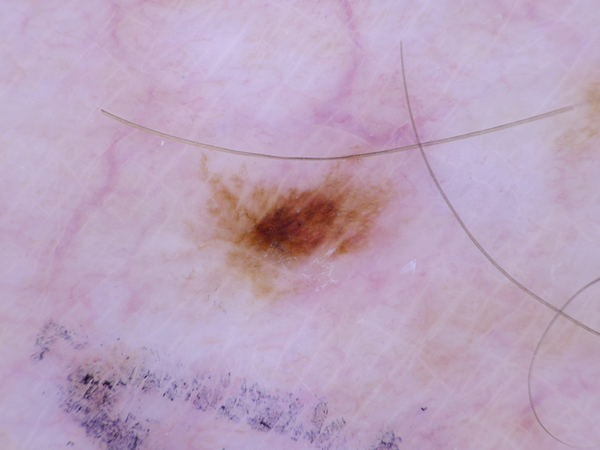} &
		\includegraphics[width=3cm,height=2.2cm]{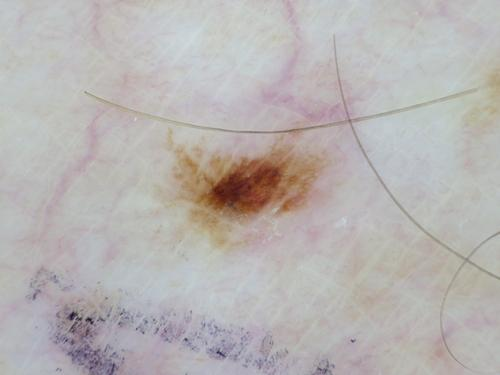}
		\\
		(a) Original images& (b) After pre-processing \\ 
	\end{tabular}     
	
	\caption[]{Examples of pre-processing by Shades of Gray algorithm \cite{finlayson2004shades}.}
	\label{fig:preprocess}
\end{figure}

\begin{figure}
	\centering
	\begin{tabular}{ccccccc}
		\includegraphics[width=1.6cm,height=1.3cm]{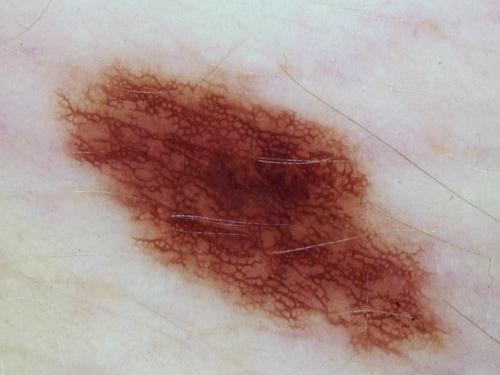} &
		\includegraphics[width=1.6cm,height=1.3cm]{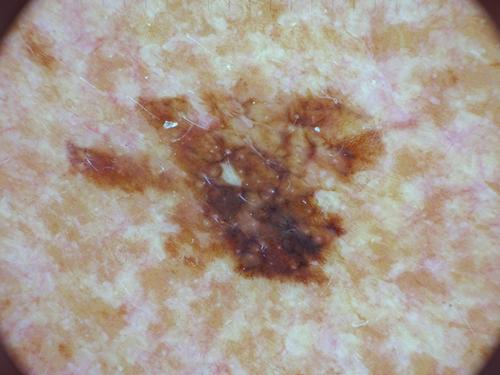} &
		\includegraphics[width=1.6cm,height=1.3cm]{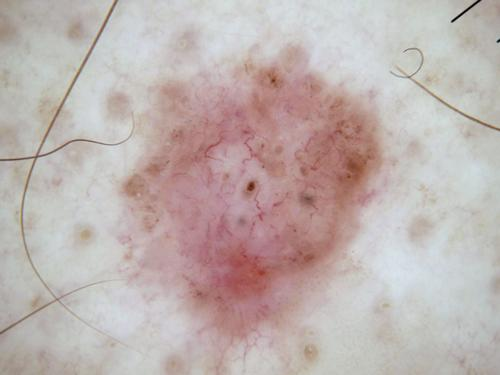} &
		\includegraphics[width=1.6cm,height=1.3cm]{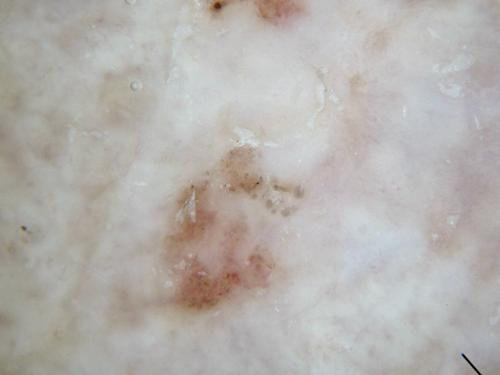} &
		\includegraphics[width=1.6cm,height=1.3cm]{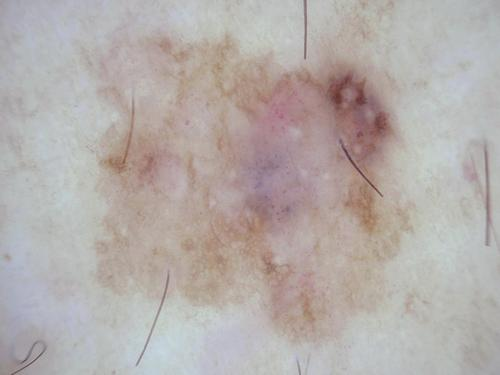} &
		\includegraphics[width=1.6cm,height=1.3cm]{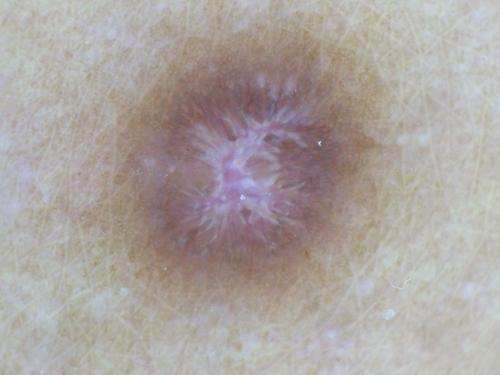} &
		\includegraphics[width=1.6cm,height=1.3cm]{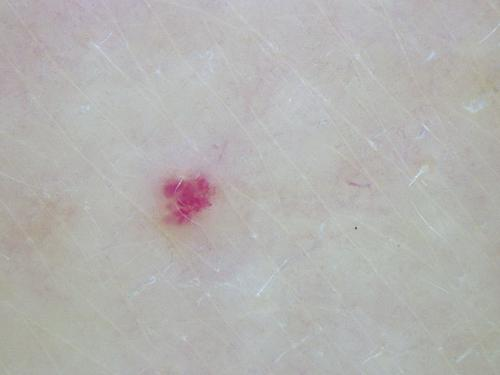}\\
		\includegraphics[width=1.6cm,height=1.3cm]{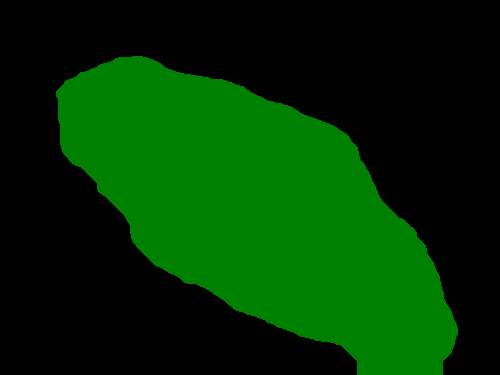} &   
		\includegraphics[width=1.6cm,height=1.3cm]{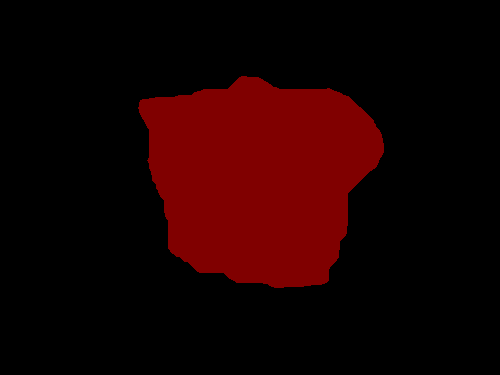}&
		\includegraphics[width=1.6cm,height=1.3cm]{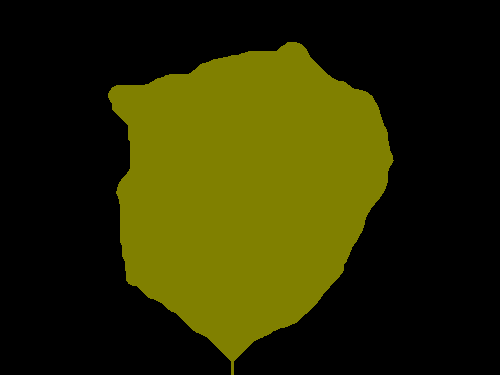}&
		\includegraphics[width=1.6cm,height=1.3cm]{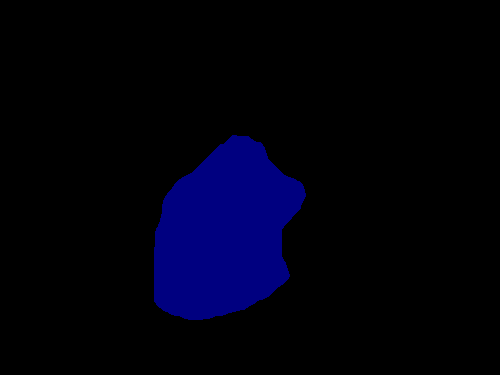}&
		\includegraphics[width=1.6cm,height=1.3cm]{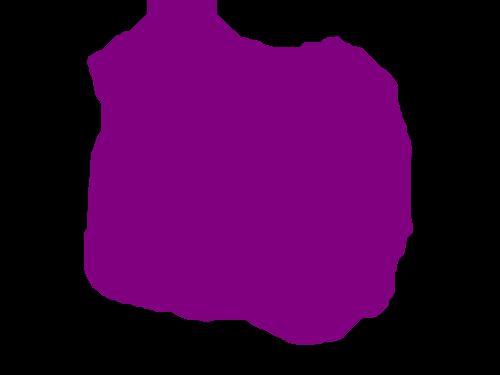} &   
		\includegraphics[width=1.6cm,height=1.3cm]{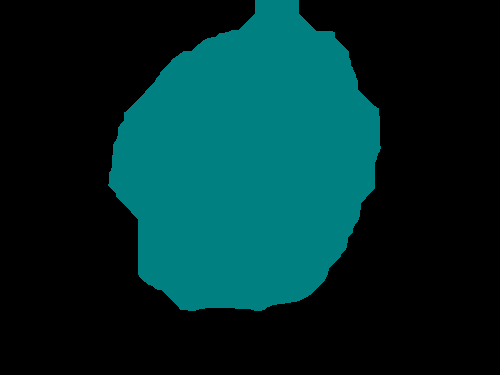}&
		\includegraphics[width=1.6cm,height=1.3cm]{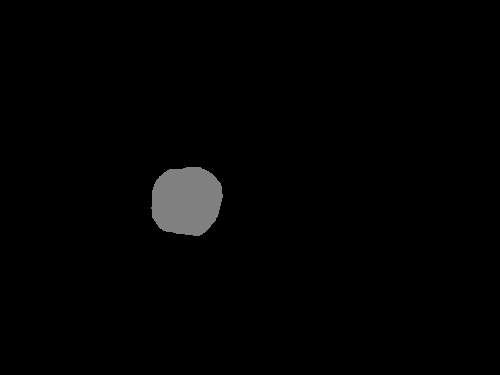}\\
		
		(a) &(b) & (c)  &(d) &(e)  & (f) & (g) \\  
	\end{tabular}     
	
	\caption[]{Illustration of seven types of skin lesion with ground truth masks where (a) Nevi (b) Melanoma (c) Basal Cell Carcinoma (d) Actinic Keratosis (e) Benign Keratosis (f) Dermatofibroma and (g) Vascular Lesion. }
	\label{fig:GT1}
\end{figure}

We used Task 1 dataset which consists of 2594 training images and segmentation masks to train DeeplabV3+ to generate the skin lesion masks for \textit{HAMM-10000 dataset}. Due to \textit{ISIC Challenge} datasets comprised of dermoscopic skin lesion taken by different dermoscope and camera devices all over the world, it is important to perform pre-processing for color normalization and illumination. We pre-process the dataset with a color constancy algorithm, Shades of Gray algorithm \cite{finlayson2004shades}, and the pre-processed results are as shown in the Fig. \ref{fig:preprocess}. With classification labels, we converted the skin lesion masks to represent different categories of skin lesions as shown in the Fig. \ref{fig:GT1}.  We used the Pascal-VOC format i.e. the input images are defined in RGB colorspace and 8-bit paletted images are used for representing the ground truth.

\subsection{DeeplabV3+}
DeepLabV3 \cite{chen2018deeplab} is one of the best performing pixel-wise classification (semantic segmentation) networks. It assigns semantic label lesion to every pixel in a dermoscopic image. DeeplabV3+ is an encoder-decoder network which make use of CNN (Xception-65) with atrous convolution layers to get the coarse score map and then, conditional random field is used to produce final output. We trained DeeplabV3+ for 100 epochs with batch size of 4, base learning learning rate of 0.0001 with learning decay factor of 0.1, decay step of 5000 and weight decay of 0.00005.  We run our experiments on a machine with the following specification: (1) Hardware: CPU - Intel i7-6700 @ 4.00Ghz, GPU - NVIDIA TITAN X 12Gb, RAM - 32GB DDR5 (2) Software: Tensor-flow.    

\begin{figure}
	\centering
	\begin{tabular}{ccc}
		\includegraphics[width=3cm,height=2.2cm]{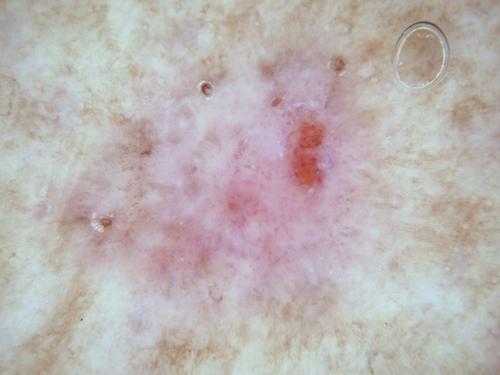} &
		\includegraphics[width=3cm,height=2.2cm]{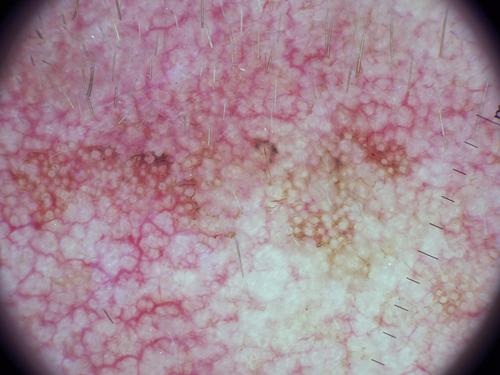} &
		\includegraphics[width=3cm,height=2.2cm]{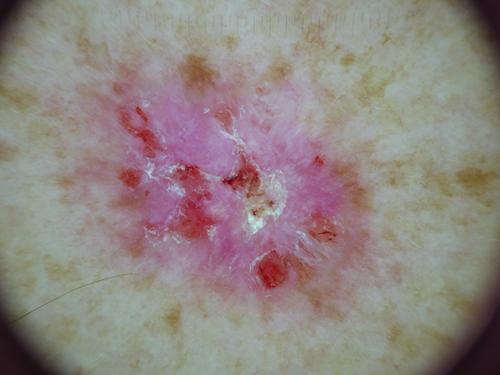} \\
		\includegraphics[width=3cm,height=2.2cm]{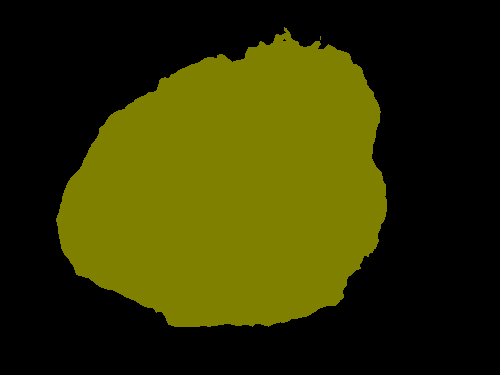} &   
		\includegraphics[width=3cm,height=2.2cm]{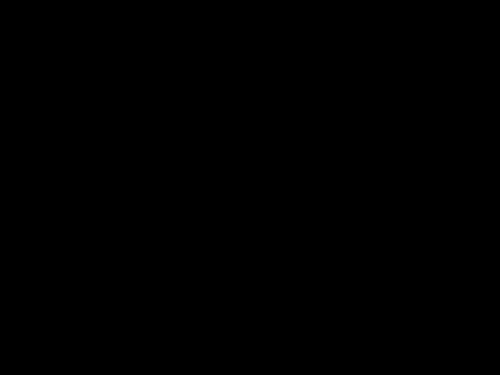}&
		\includegraphics[width=3cm,height=2.2cm]{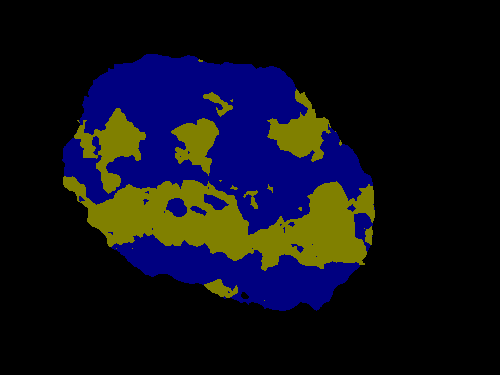}\\
		
		(a) Single Detection & (b) No Detection  & (c) Multiple Detection  \\ 
	\end{tabular}     
	
	\caption[]{Examples of different types of semantic segmentation for validation set: (a) result with one class lesion type; (b) result with no lesion detected; and (c) result with multiple lesion types. }
	\label{fig:det1}
\end{figure}

\subsubsection{Post-processing Methods to determine lesion class}
\begin{figure}
			\centering
			\includegraphics[scale=0.7]{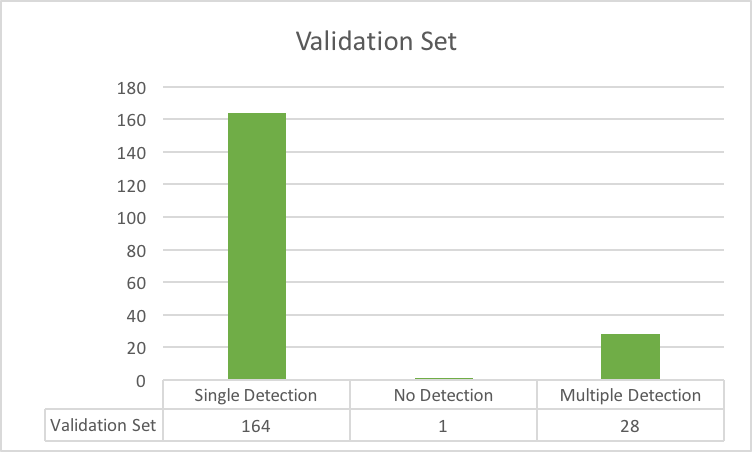}
			\caption{Number of each type of detection in the validation set.}
			\label{fig:det2}
		\end{figure}

After training, we used the images in validation set and testing set to produce outputs. We received three types of results from the inference as shown in the Fig. \ref{fig:det1}. The number of lesion types detected is shown in  Fig. \ref{fig:det2}.
		
For single type detection, we directly assume diagnosis confidence of 1.0 of detected lesion class. There are very few cases of no detection in both validation set and testing set, we assumed it as 1.0 for Nevi class. For multi-class detection, we adopted 2 different strategies to determine class of lesion. First is diagnosis confidence based on number of pixels of each lesion class in the detection means that lesion class which has most pixels in the detection is considered. Second strategy is priority based on the number of images of different type of skin lesion in the dataset as shown in the Table \ref{tab:tradFse}. Third method is again an priority based strategy with preference of the malignant lesions over the benign and number of images in the training set according to the Table \ref{tab:tradFseddd}.   

	\begin{table}[]
		\centering
		\small\addtolength{\tabcolsep}{2pt}
		\renewcommand{\arraystretch}{2}
		\caption{Priority strategy based on distrubution of skin lesions in HAMM training set}
		\label{tab:tradFse}
		\scalebox{0.78}{
			\begin{tabular}{llc}
				\hline
				Priority & Lesion Class & Number of Images in training set   \\ \hline\hline
				1  & Dermatofibroma       & 115                   \\
				2 & Vascular Lesion           & 142                           \\
				3  & Actinic Keratosis         & 327                          \\
				4 & Basal Cell Carcinoma        & 514                   \\ 
				5  & Benign Keratosis       & 1099                         \\
				6 & Melanoma         & 1113                     \\
				7      & Nevi         & 6705                     \\	\hline
			\end{tabular}}
		\end{table}

		\begin{table}[]
			\centering
			\small\addtolength{\tabcolsep}{2pt}
			\renewcommand{\arraystretch}{2}
			\caption{Priority strategy based on benign or maligant and number of images in HAMM training set}
			\label{tab:tradFseddd}
			\scalebox{0.78}{
				\begin{tabular}{lllc}
					\hline
					Priority & Lesion Class & Benign or Malignant & Number of Images in training set   \\ \hline\hline
					1 & Dermatofibroma       & Malignant & 115\\ 
					2 & Basal Cell Carcinoma  & Malignant & 514                   \\
					3 & Melanoma         & Malignant & 1113 \\
					4 & Vascular Lesion    & May be Malignant or Benign & 142                           \\
					5  & Actinic Keratosis         & Malignant potential &327                          \\
					5  & Benign Keratosis   & Benign    & 1099                         \\
					7      & Nevi         & Benign & 6705                     \\	\hline
				\end{tabular}}
			\end{table}
			
			%https://onlinelibrary.wiley.com/doi/full/10.1111/j.1524-4725.2008.34272.x Dermatofibroma
%The outputs produced by both methods have very high \textit{Jaccard Similarity Index (JSI)} on the validation set. For each input image, \textit{JSI} score between the outputs of both network is evaluated, if it is less than our set threshold value, then larger output is chosen over smaller output. The threshold value is chosen between 0.0 to 0.50 for each method depends on the best validation score. For the validation phase, Ensemble-Mask received 75.92, Ensemble-Deep recieved 76.67 and Ensmeble-Compare received 79.47. Smaller \textit{JSI} score between output is a good indication if one method could not produce any inference or covers very smaller area of actual skin lesion in the testing image. Our ensemble methods relies on the \textit{JSI} score according to the following equations.

\section{Validation and Conclusion} 
For lesion diagnosis challenge, evaluation metric called \textit{normalized multi-class accuracy} is used to determine the winner of the competition. We did not report Validation scores due to the rules of competition. We will report our complete analysis and final test results in a later stage as it will be announced later this month. We are submitting these three methods for this year's competition.   

\newpage
\addtocmark[2]{Author Index} % additional numbered TOC entry
\renewcommand{\indexname}{Author Index}
\printindex

\bibliographystyle{unsrt}
\bibliography{skin}

\end{document}